# Tail Sensitivity Analysis in Bayesian Networks

**Enrique F. Castillo**
Dept. of Applied Mathematics
and Computational Sciences.
University of Cantabria
39005 Santander, Spain
e-mail:castie@ccaix3.unican.es

**Cristina Solares**
Dept. of Applied Mathematics
and Computational Sciences.
University of Cantabria
39005 Santander, Spain
e-mail:solaresc@ccaix3.unican.es

**Patricia Gómez**
Dept. of Applied Mathematics
and Computational Sciences.
University of Cantabria
39005 Santander, Spain

## Abstract

The paper presents an efficient method for simulating the tails of a target variable $Z = h(\mathbf{X})$ which depends on a set of basic variables $\mathbf{X} = (X_1, \dots, X_n)$. To this aim, variables $X_i; i = 1, \dots, n$ are sequentially simulated in such a manner that $Z = h(x_1, \dots, x_{i-1}, X_i, \dots, X_n)$ is guaranteed to be in the tail of $Z$. When this method is difficult to apply, an alternative method is proposed, which leads to a low rejection proportion of sample values, when compared with the Monte Carlo method. Both methods are shown to be very useful to perform a sensitivity analysis of Bayesian networks, when very large confidence intervals for the marginal/conditional probabilities are required, as in reliability or risk analysis. The methods are shown to behave best when all scores coincide. The required modifications for this to occur are discussed. The methods are illustrated with several examples and one example of application to a real case is used to illustrate the whole process.

## 1 INTRODUCTION

Let $\mathbf{X} = \{X_1, X_2, \dots, X_n\}$ be a set of $n$ discrete variables, taking values in the set $\{0, 1, \dots, r_i\}$. A Bayesian network over $X$ is a pair $(D, P)$, where the graph $D$ is a directed acyclic graph (DAG) over $\mathbf{X}$ and $P = \{P_1(x_1|\pi_1), \dots, P_n(x_n|\pi_n)\}$ is a set of $n$ conditional probabilities, one for each variable, where $\Pi_i$ is the set of parents of node $X_i$. If we denote

$$\theta_{ij\pi} = P_i(X_i = j|\Pi_i = \pi), \; j \in \{0, \dots, r_i\}, \quad (1)$$

where $\pi$ is any possible instantiation of the parents of $X_i$, the joint probability density of $X$ can be considered as a $\theta$-parametric family and written as:

$$P(x_1, x_2, \dots, x_n) = \prod_{i=1}^{n} P_i(x_i|\pi_i). \quad (2)$$

Castillo, Gutiérrez and Hadi [3, 4] have shown that the marginal $P(X_i = j)$, and conditional probabilities $P(X_i = j|E = e)$, where $E$ is a set of evidential nodes with known values $e$, are polynomials and quotients of polynomials, respectively of the $\theta$-parameters which are first degree in each parameter.

A sensibility analysis of a Bayesian network consists of assuming the $\theta$ parameters as random variables, instead of constant values, and calculating the density functions of the marginal/conditional probabilities above, which are monotone functions of the parameters. In some cases, as in reliability or risk analysis very high confidence intervals are needed, which implies the estimation of their tails. In this paper we deal with a more general problem which consists of estimating the tail of a random variable which is related to other basic variables by a monotone relation. We assume that the cdf of the target variable is not directly available but determined through the basic variables.

Monte Carlo simulation allows dealing with a random variable which is related to other variables by a complex relation. The method performs well in the central part of the distribution, but gives very poor approximations in the tails, as, for example, the estimation of small or large percentiles. In Engineering design only tails are important. The engineer is only interested in the occurrences of either very large values of magnitudes (temperatures, winds, waves, earthquakes, etc.) or very low values of the same magnitudes, because they produce structural, supply or environmental problems. This has motivated the appearance of extreme value theory (see Galambos [6] or Castillo [2]) and several papers deal with the estimation of large percentiles (see Weissman [9]).

Several simulation methods have been proposed for simulating random samples in Bayesian networks, as: stochastic simulation (Pearl [7]), likelihood weighing (Shachter and Peot [8]; hybrid methods of logic sampling and stochastic simulation (Chavez and Cooper [5]), stratified sampling (Bouckaert, Castillo and Gutiérrez [1], Castillo, Gutiérrez and Hadi [4]), etc. However, these methods have not been applied to solve the tail estimation problem.

We note here that estimation of extreme percentiles is difficult from real samples but if we can control the simulation method the thing is completely different. In this paper we present a method which allows simulating the tails of the target variable with a reduced or null proportion of rejections. This means that each data point in the simulated sample belongs to the desired tail with a high probability.

The paper is structured as follows. In Section 2 we describe how the methods can be applied to several cases of monotone functions and we compare the simulations with the exact tails. In Section 2.1 we give a method which allows simulating the tails directly, that is, the simulated sample values belong to the tail. We also give the scores $w$ we need to consider for simulating with a distribution which does not coincide with the real population. In Section 2.2 we show how this method can be improved by using simulation procedures with equal scores. In Section 2.3 we describe an alternative method which leads to a relatively small rejection proportion. In Section 3 we present one application of Bayesian networks to a real example. Finally, Section 4 gives some conclusions.

# 2 THE PROPOSED METHODS

The main idea of the proposed method consists of simulating only the tail of the target variable. Assume that $Z = h(\mathbf{X})$, that is, we have a random variable $Z$ related to a set $\mathbf{X} = \{X_1, \ldots, X_n\}$ of basic random variables by an increasing (decreasing) relation.

Note that if $h(\mathbf{X})$ is decreasing, we can work with $-Z$ instead of $Z$. Assume also that $\infty < a_i \leq X_i \leq b_i < \infty$, i.e., the random variables $X_i; i = 1, \ldots, n$ are bounded. Then, $h(\mathbf{a}) \leq Z \leq h(\mathbf{b})$, that is, $h(\mathbf{a})$ and $h(\mathbf{b})$ are lower and upper bounds of $Z$, respectively.

To illustrate the proposed methods we use a very simple bidimensional example. Let $h(x_1, x_2) = 1 - x_1x_2$; $0 \leq x_1, x_2 \leq 1$. The minimum of $h(x_1, x_2)$ is 0, which is attained at the point $(1, 1)$ (see Figure 1). If we are interested in the left tail $\mathcal{T} = \{0 \leq h(x_1, x_2) \leq \epsilon\}$ of $h()$, we want to simulate the $A$ dark shadowed region in Figure 1. We have several options as:

1. **The Standard Monte Carlo Method** : We simulate the random values in the unit square and reject the sample if it outside the $\mathcal{T}$ region. It leads to a very high rejection proportion.
2. **The No Rejection Method** : We simulate samples in the $A$ region directly. It is very efficient, since the rejection proportion is null, but can be difficult to implement because of the complexity of the tail boundary.
3. **The Reduced Rejection Method** : We simulate points in an intermediate region as the region limited by the tangent hyperplane to $\mathcal{T}$ (see Figure). It is easy to apply (we invert linear equations), and convenient (low rejection proportion).

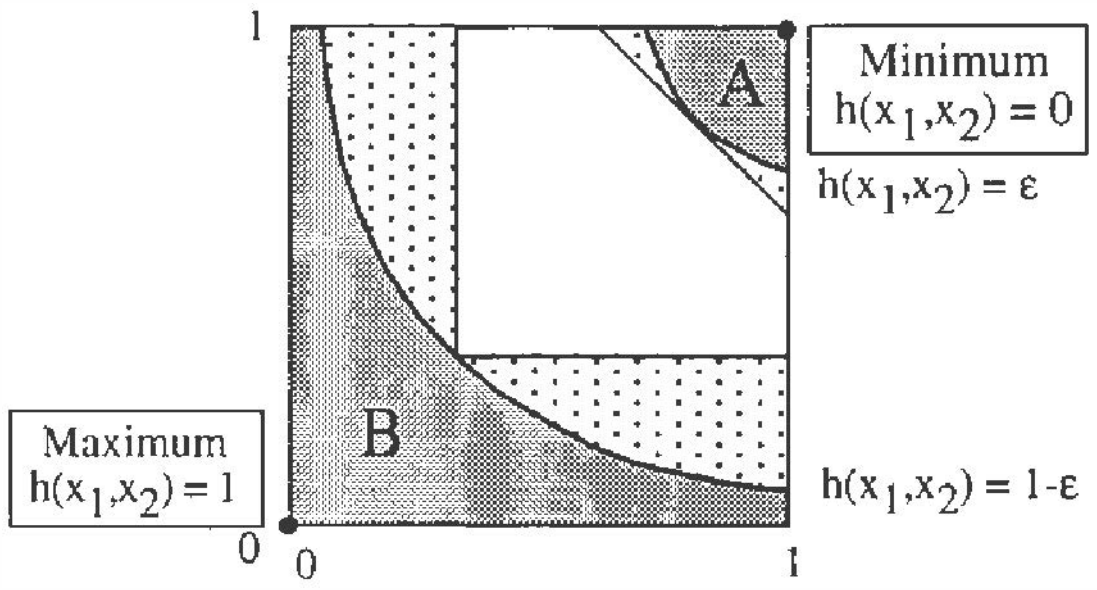

Figure 1: Illustration of different possibilities for simulating the right and left tails.

The rejection proportion is

$$p_\epsilon = \sum_{x \in \mathcal{S}_\epsilon - \mathcal{T}_\epsilon} P_S(x) / \sum_{x \in \mathcal{T}_\epsilon} P_S(x), \quad (3)$$

where $S_\epsilon$ is the simulated region for a given $\epsilon$. It is convenient to have $\lim_{\epsilon \to 0} p_\epsilon << 1$.

For the no rejection method we have $S_\epsilon = \mathcal{T}_\epsilon$ and then $p_\epsilon = 0$. On the contrary, for the standard Monte Carlo method $S_\epsilon = S$ and $\mathcal{T}_\epsilon \to \emptyset$; thus, we have $\lim_{\epsilon \to 0} p_\epsilon = 1$.

## 2.1 THE NO REJECTION METHOD

We describe this method only for the left tail.

We aim to approximate the cdf in the left tail $\mathcal{T} = \{z : h(\mathbf{a}) < z \leq h(\mathbf{a}) + \epsilon\}$ of $Z$, by simulating the random variable $\mathbf{X}$ restricted to $Z \leq h(\mathbf{a}) + \epsilon$.

For any $X_i$, let us denote $h_i^{-1}(\mathbf{x}_{i-1}, z, \mathbf{x}^{i+1})$ the inverse of $h(\mathbf{x})$ with respect to $x_i$, where we have denoted $\mathbf{x}_i = (x_1, \ldots, x_i)$ and $\mathbf{x}^i = (x_i, \ldots, x_n)$. The proposed method sequentially simulates variables $\mathbf{X}$ in the following form. Assume that we have already simulated variables $X_1 = x_1, \ldots, X_{i-1} = x_{i-1}$. Then we simulate $X_i$ such that

$$h(\mathbf{x}_{i-1}, \mathbf{a}^i) < h(\mathbf{x}_{i-1}, X_i, \mathbf{X}^{i+1}) \leq h(\mathbf{a}) + \epsilon \quad (4)$$

Note that once variables $\mathbf{X}_{i-1}$ have been simulated, the new lower bound of $Z$ is $h(\mathbf{x}_{i-1}, \mathbf{a}^i)$.

From (4) we get

$$\ell_i(\mathbf{X}^{i+1}) < X_i \leq u_i(\mathbf{X}^{i+1}) \quad (5)$$

where

$$\ell_i(\mathbf{X}^{i+1}) = h_i^{-1}(\mathbf{x}_{i-1}, h(\mathbf{x}_{i-1}, \mathbf{a}^i), \mathbf{X}^{i+1}) \quad (6)$$

$$u_i(\mathbf{X}^{i+1}) = h_i^{-1}(\mathbf{x}_{i-1}, h(\mathbf{a}) + \epsilon, \mathbf{X}^{i+1}) \quad (7)$$

Since $\mathbf{X}^{i+1}$ have not been simulated yet, we must choose the largest possible interval which, taking into account the constraint $a_i < X_i \leq b_i$, is

$$L_i = max \left[ \min_{\mathbf{X}^{i+1}} \ell_i(\mathbf{X}^{i+1}), a_i \right] < X_i \leq min \left[ \max_{\mathbf{X}^{i+1}} u_i(\mathbf{X}^{i+1}), b_i \right] = U_i \quad (8)$$

Once $L_i$ and $U_i$ are known, we can simulate $X_i$, with density proportional to $f(x_i|\mathbf{x}_{i-1})$ in the region $L_i < X_i \leq U_i$. Note that, since all sample values belong to the target region, no rejection occurs.

### 2.1.1 SIMULATION ALGORITHM

Thus,we have the following algorithm:

**Algorithm 1**

**INPUT**

- *An increasing function defining the target variable:* $Z = h(\mathbf{X})$
- *A set of n conditional probabilities* $f(X_i|\mathbf{X}_{i-1})$.
- *Lower and upper bounds of the basic random variables* $\mathbf{X}$*:* $\mathbf{a}$ *and* $\mathbf{b}$.
- *Sample size m and desired departure* $\epsilon$ *from lower* $h(\mathbf{a})$ *or upper bound* $h(\mathbf{b})$ *of the target random variable Z.*

**OUTPUT**

- *A sample of size m from the left tail* $\mathcal{T} = \{z : h(\mathbf{a}) < z \leq h(\mathbf{a}) + \epsilon\}$ *or the right tail* $\mathcal{T} = \{z : h(\mathbf{b}) - \epsilon < z \leq h(\mathbf{b})\}$ *of the target Z.*

**STEPS**

- **Step 1** *: Simulate sequentially* $X_i; i = 1, \ldots, n$ *in the interval* $L_i < X_i \leq U_i$*, using* $f(X_i|\mathbf{x}_{i-1})$*, that is, we simulate truncated variables, where* $x_i; i = 1, \ldots, n$ *are the simulated values.*
- **Step 2** *: Calculate the simulated sample value* $z_j = h(\mathbf{x})$ *and assign it the score*

$$w_j = \prod_{i=1}^{n} \left[F_{X_i|\mathbf{X}_{i-1}}(U_i|\mathbf{x}_{i-1}) - F_{X_i|\mathbf{X}_{i-1}}(L_i|\mathbf{x}_{i-1})\right].$$

- **Step 3** *: Store the pair* $(z_j, w_j)$.
- **Step 4** *: Repeat steps 1 to 3 m times.*
- **Step 5** *: Sort the pairs* $(z_j, w_j); j = 1, \ldots, m$ *with respect to* $z_j$.
- **Step 6** *: Replace* $w_j$ *in the pairs* $(z_j, w_j); j = 1, \ldots, m$ *by* $\sum_{k=1}^{j} w_k/m$*, for the left tail and by* $1 - \sum_{k=j+1}^{m} w_k/m$*, for the right tail.*
- **Step 7** *: The resulting* $w_j; j = 1, \ldots, m$ *are the simulated approximations to* $F_Z(z_j)$.

The validity of this method relies on the well known rejection method. Note that here the sampling distribution is

$$P_S(x_i) = \frac{f(x_i|\mathbf{x}_{i-1})}{F_{X_i|\mathbf{X}_{i-1}}(U_i|\mathbf{x}_{i-1}) - F_{(X_i|\mathbf{X}_{i-1})}(L_i|\mathbf{x}_{i-1})} \quad (9)$$

and then the score becomes:

$$w = \prod_{i=1}^{n} \left[F_{X_i|\mathbf{X}_{i-1}}(U_i|\mathbf{x}_{i-1}) - F_{X_i|\mathbf{X}_{i-1}}(L_i|\mathbf{x}_{i-1})\right]. \quad (10)$$

**Example 1** *[Right tail of sums]* In the case of sums $h(\mathbf{X}) = \sum_{i=1}^{n} X_i$, and the functions $\ell_i(\mathbf{X}^{i+1})$ and $u_i(\mathbf{X}^{i+1})$ in (6) and (7) become

$$\begin{aligned} \ell_i(\mathbf{X}^{i+1}) &= -\sum_{k=1}^{i-1} x_k + \sum_{k=1}^{n} b_k - \epsilon - \sum_{k=i+1}^{n} X_k. \\ u_i(\mathbf{X}^{i+1}) &= \sum_{k=i}^{n} b_k - \sum_{k=i+1}^{n} X_k. \end{aligned} \quad (11)$$

Thus, from (8) we get

$$L_i = max(\sum_{k=1}^{i} b_k - \sum_{k=1}^{i-1} x_k - \epsilon, a_i); \quad U_i = b_i. \quad (12)$$

The exact cdf of the sum of $n$ uniforms is given by

$$F(x) = \frac{1}{n!} \sum_{r=0}^{\lfloor x \rfloor} \binom{n}{r} (t - r)^n; 0 < x < n \quad (13)$$

where $\lfloor x \rfloor$ is the integer part of $x$.

Figure 2 gives the exact and the simulated tails of the sum of four uniforms $U(0, 1)$, respectively, for a sample size $n = 1000$, when the tail interval $(3.88, 4.0)$.

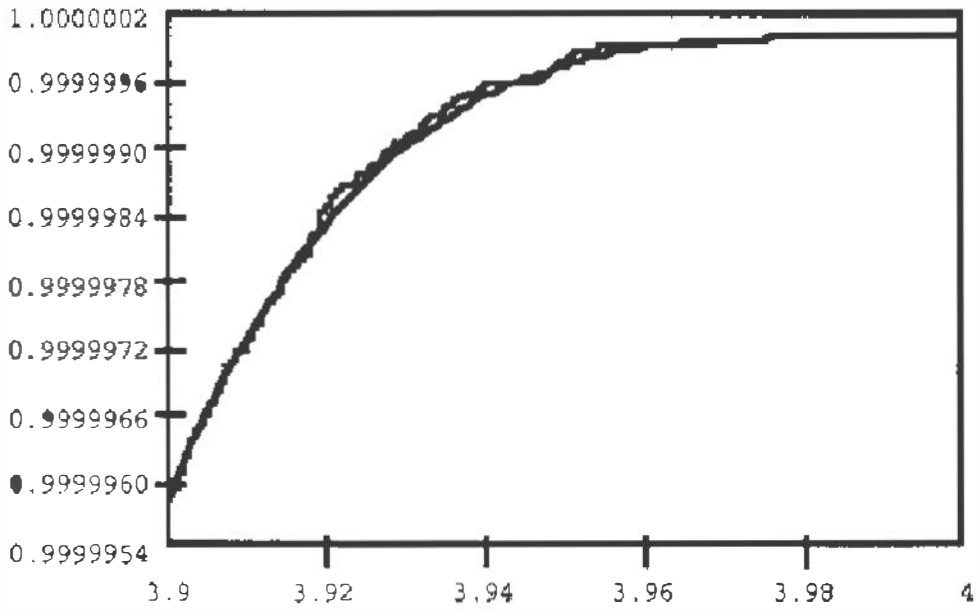

Figure 2: Exact and simulated tail $(3.88, 4.0)$ of the sum of four uniforms $U(0, 1)$.

**Example 2** *[Left tail of products]* In the case of products of non-negative variables $h(\mathbf{X}) = \prod_{i=1}^{n} X_i$, and the functions $\ell_i(\mathbf{X}^{i+1})$ and $u_i(\mathbf{X}^{i+1})$ in (6) and (7) become

$$\ell_i(\mathbf{X}^{i+1}) = \frac{\prod_{k=i}^{n} a_k}{(\prod_{k=i+1}^{n} X_k)}. \quad (14)$$

$$u_i(\mathbf{X}^{i+1}) = \frac{\prod_{k=1}^{n} a_k + \epsilon}{\prod_{k=1}^{i-1} x_k \prod_{k=i+1}^{n} X_k}. \tag{15}$$

Thus, $L_i$ and $U_i$ become (see (8))

$$L_i = max\left( \min_{\mathbf{X}^{i+1}} \frac{\prod_{k=i}^{n} a_k}{\prod_{k=i+1}^{n} X_k}, a_i \right) = a_i \tag{16}$$

$$U_i = min\left( \frac{\prod_{k=1}^{n} a_k + \epsilon}{\prod_{k=1}^{i-1} x_k \prod_{k=i+1}^{n} a_k}, b_i \right) \tag{17}$$

The exact cumulative distribution function of the product of $n$ uniforms $U(0,1)$ is

$$F(x) = x\left( \sum_{i=0}^{n-1} \frac{(-\log x)^i}{i!} \right); \; 0 \le x \le 1. \tag{18}$$

Figure 3 shows the simulated left tail of $x_1x_2x_3$ with the no rejection method for $\epsilon = 0.001$.

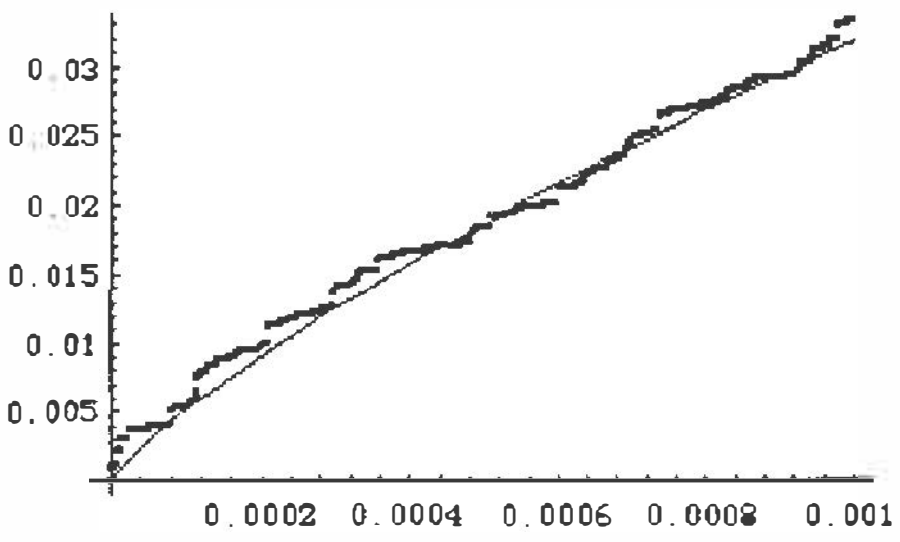

Figure 3: Simulated left tail of $x_1x_2x_3$ with the no rejection method for $\epsilon = 0.001$.

## 2.2 IMPROVING THE METHOD

The no rejection method performs very well for small number of basic variables but deteriorates if this number increases. The main reason for this is that the number of feasible different instantiations blows up with the number of basic variables and the associated scores become very far appart. This problem can be solved if we design simulating procedures leading to similar scores for all instantiations, the ideal situation being all scores to be equal. Now we show how this optimal solution can be theoretically achieved.

Our simulation density and scores are (9) and (10). However, if we choose

$$P_S(x_i) = \frac{f(x_i|\mathbf{x}_{i-1})[G_{i+1}(U_{i+1}) - G_{i+1}(L_{i+1})]}{G_i(U_i) - G_i(L_i)}$$
$$P_S(x_n) = \frac{f(x_n|\mathbf{x}_{n-1})}{G_n(U_n) - G_n(L_n)} \tag{19}$$

where $G_i(x_i)$ is the cdf associated with the pdf

$$g_i(x) \doteq f(x|\mathbf{x}_{i-1})[G_{i+1}(U_{i+1}) - G_{i+1}(L_{i+1})];$$
$$g_n(x) = f(x|\mathbf{x}_{n-1}), with\ L_i \le x \le U_i; \; i = 1, \ldots, n \tag{20}$$

the score becomes $w = G_1(U_1) - G_1(L_1)$, which is independent of the sample and implies constant scores.

## 2.3 REDUCED REJECTION METHOD

The main problem with the proposed method consists of determining the inverse functions $h_i(.)$. This problem does not exist for the polynomial or rational functions of the marginal (conditional) probabilities, which need not be inverted but calculated from two (three) evaluations of $h()$. However, to avoid this problem in other cases, we can modify the previous method but we have to pay with some rejections. The central idea consists of simulating an alternative region containing the tail region $\mathcal{T}$. For example, if we want to simulate the tail associated with region $A$ in Figure 1, we simulate the region limited by the hyperplane tangent to $A$ which is parallel to the tangent hyperplane to the function $h(\mathbf{x}) = h(\mathbf{a})$ at the point $\mathbf{a}$, that is, the dotted region in the upper-right corner of Figure 1. If the sample point is outside $A$, we reject it, otherwise we accept it.

To this end, we obtain the point of tangency $(\mathbf{x}_0)$, which coordinates are the solutions of the system of equations

$$h(\mathbf{x}_0) = h(\mathbf{a}) + \epsilon \tag{21}$$

$$\frac{\left.\dfrac{\partial h(\mathbf{x})}{\partial x_i}\right|_{(\mathbf{x}_0)}}{\left.\dfrac{\partial h(\mathbf{x})}{\partial x_n}\right|_{(\mathbf{x}_0)}} = \frac{\left.\dfrac{\partial h(\mathbf{x})}{\partial x_i}\right|_{(\mathbf{a})}}{\left.\dfrac{\partial h(\mathbf{x})}{\partial x_n}\right|_{(\mathbf{a})}}; \; i = 1, \ldots, n-1 \tag{22}$$

Equation (21) forces the point $(\mathbf{x}_0)$ to belong to the surface $h(\mathbf{x}) = h(\mathbf{a}) + \epsilon$ and Equation (22) forces the tangent hyperplane to $h(\mathbf{x}) = h(\mathbf{a}) + \epsilon$ at $(\mathbf{x}_0)$ to be parallel to the tangent hyperplane to $h(\mathbf{x}) = h(\mathbf{a})$ at the point $\mathbf{a}$.

Thus, we get the hyperplane and region

$$\sum_{k=1}^{n} (x_k - x_{k0}) \left.\frac{\partial h(\mathbf{x})}{\partial x_k}\right|_{(\mathbf{a})} = 0 \tag{23}$$

$$\sum_{k=1}^{n} (x_k - x_{k0}) \left.\frac{\partial h(\mathbf{x})}{\partial x_k}\right|_{(\mathbf{a})} \gtrless 0 \tag{24}$$

which we initially simulate. The sign ($>$ or $<$) to be used in (24) depends on the tail we are dealing with. For this method to be valid, the quadratic form

$$\sum_{i=1}^{n}\sum_{j=1}^{n} \frac{\partial^2 h(\mathbf{a})}{\partial x_i \partial x_j}(x_i - a_i)(x_j - a_j), \tag{25}$$

must be negative definite. This guarantees that the target tail $\mathcal{T}$ is inside the simulated region for small values of $\epsilon$.

In some cases, this method fails, since the region $A$ is not all on the same side of the hyperplane. This happens, for example, in the case of the left tail of $h(x_1, x_2)$ (see Figure 1). In this case we can use other regions as the one indicated (dotted in the Figure 1). This region corresponds to the equation

$$min\,(x_1, x_2) < \sqrt{\epsilon}. \tag{26}$$

Thus, Step 1 in Algorithm 1 transforms to:

- **Step 1** : Simulate sequentially $X_i; i = 1, \dots, n$ in the interval
$$L_i < X_i \leq U_i, \quad (27)$$
using $f(X_i|\mathbf{x}_{i-1})$, where now the interval $(L_j, U_j)$ is calculated as follows. Assume that we have already simulated $X_1 = x_1, \dots, X_{i-1} = x_{i-1}$, and we want to simulate $X_i$, then, one of the bounds for $X_i$ is $a_i$ and we can calculate the other bound for $X_i$ assuming $X_j = a_j;\ j = i+1, \dots, n$ and forcing the simulated point to belong to the hyperplane (23), that is,
$$\sum_{k=1}^{n}(X_k - x_{k0})\frac{\partial h(\mathbf{a})}{\partial x_k} = 0, \quad (28)$$
from which we get the $b_i$ bound
$$x_{i0} - \frac{\sum_{k=1}^{i-1}(x_k - x_{k0})\frac{\partial h(\mathbf{a})}{\partial x_k} + \sum_{k=i+1}^{n}(a_k - x_{k0})\frac{\partial h(\mathbf{a})}{\partial x_k}}{\frac{\partial h(\mathbf{a})}{\partial x_i}} \quad (29)$$
which leads to
$$L_j = a_j;\ U_j = b_j \ if\ b_j \geq a_j \quad (30)$$
$$L_j = b_j;\ U_j = a_j \ if\ b_j < a_j \quad (31)$$
Next, we check whether $h(\mathbf{x}) \leq h(\mathbf{a}) + \epsilon$, and if it is we accept the sample point, otherwise, we reject it.

**Example 3** *[Right tail of three uniforms $U(0,1)$]* In this example we use the method described above to simulate the right tail of
$$h(x_1, x_2, x_3) = x_1 x_2 x_3,\ x_1, x_2, x_3 \sim U[0,1], \quad (32)$$
that is, to the product of three uniforms $U(0,1)$. The maximum is attained at the point $(a_1, a_2, a_3) = (1,1,1)$ and the system (21)-(22) becomes
$$x_{10}x_{20}x_{30} = 1 - \epsilon \quad (33)$$
$$\frac{x_{20}x_{30}}{x_{10}x_{20}} = \frac{x_{10}x_{30}}{x_{10}x_{20}} = 1 \quad (34)$$
from which we get $x_{10} = x_{20} = x_{30} = \sqrt[3]{1-\epsilon}$ and the simulation region becomes
$$\sum_{k=1}^{3}\left(x_k - \sqrt[3]{1-\epsilon}\right) > 0 \quad (35)$$
From Equation (29), the bound $b_i$ becomes
$$b_i = i - n\left[1 - \sqrt[3]{1-\epsilon}\right] - \sum_{k=1}^{i-1} x_k \quad (36)$$
Thus, we simulate $X_i; i = 1,2,3$ in the interval $(L_i, U_i)$, where $L_i$ and $U_i$ are given by (30) and (31), respectively. Next we calculate $h(x_1, x_2, x_3) = x_1 x_2 x_3$ and check whether or not it is larger than $1 - \epsilon$. If it is, we calculate the corresponding score $w_i = \prod_{j=1}^{3}(U_j - L_j)/m$, where $m$ is the total sample size, add one unit to the effective sample size counter $m^*$ and accept the simulated value; otherwise we reject it. Finally, we acumulate scores in the effective sample and calculate the rejection proportion as $r = (m - m^*)/m$. Taking into account that
$$Vol[T_\epsilon] = 1-(\epsilon-1)(\log(1-\epsilon)-1)+[(\epsilon-1)\log(1-\epsilon)^2]/2 \quad (37)$$
and
$$Vol[S_\epsilon] = \frac{3^3\left(1-\sqrt[3]{1-\epsilon}\right)^3}{6} \quad (38)$$
where $Vol[T_\epsilon]$ and $Vol[S_\epsilon]$ are the volumes associated with the tail $T_\epsilon$ and the simulation $S_\epsilon$ regions, respectively, the rejection proportion becomes
$$p_\epsilon = \frac{Vol[S_\epsilon] - Vol[T_\epsilon]}{Vol[S_\epsilon]} \quad (39)$$
which for small values of $\epsilon$ becomes
$$p_\epsilon = \frac{\epsilon}{4} + \frac{4\epsilon^2}{45} \quad (40)$$
thus, the asymptotic rejection proportion is 0.

Figure 4 shows a simulated right tail for $\epsilon = 0.1$ with rejection proportion 0.203.

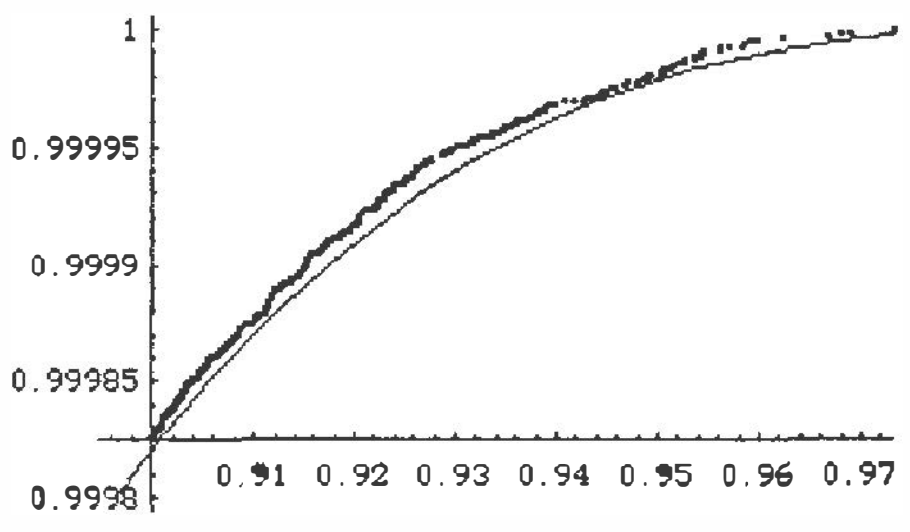

Figure 4: Simulated right tail of $x_1, x_2, x_3$ with a rejection proportion equal to 0.203.

For the left tail we simulate in the region $min(x_1, x_2, x_3) < \sqrt[3]{\epsilon}$. To this end, we simulate $X_i, i = 1, \dots, n-1$ in the interval $[0,1]$ and $X_n$ in the same interval if some of the previous $X_i$ are smaller than $\sqrt[3]{\epsilon}$ and, in the interval $[0, \sqrt[3]{\epsilon}]$, otherwise.

Figure 5 shows the simulated cdf for $\epsilon = 0.01$. The rejection proportion was 0.727.

In this case we have
$$Vol[T_\epsilon] = \frac{\epsilon(2 - 2\log\epsilon + (\log\epsilon)^2}{2} \quad (41)$$
and
$$Vol[S_\epsilon] = 1 - \left(1 - \sqrt[3]{\epsilon}\right)^3 \quad (42)$$
and the asymptotic rejection proportion is 1. Thus, the simulation region must be improved if smaller percentiles are desired.

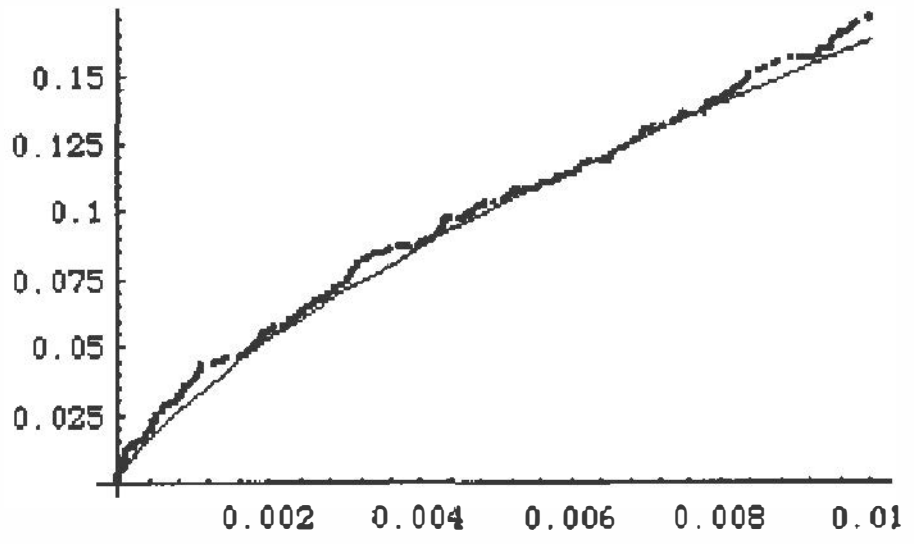

Figure 5: Simulated left tail of $x_1x_2x_3$ with a rejection proportion equal to 0.727.

### 2.4 PERFORMANCE OF THE METHOD

When the proposed method is compared with the standard Monte Carlo simulation method, we can conclude the following:

1. The whole sample or a reasonable proportion of the sample values are in the selected tail interval for the proposed method, while only a very small fraction of the sample or no sample point is in it for the standard Monte Carlo method.
2. Percentiles corresponding to very low or high probabilities are very closely estimated with a sample of relatively small size. For this case the Monte Carlo method gives very bad estimates or is enable to give one.
3. The method seems to work well for any percentile (see Figures).
4. The no rejection method deteriorates for increasing number of basic variables.

## 3 AN EXAMPLE

In this section we show an example of application to probability risk assessment. This is one of the fields where the presented method fits very well, since tails play the most important role.

Figure 6 shows the simplified flow diagram of a typical standby system and Table 1 gives the notation and the physical meaning of all involved variables.

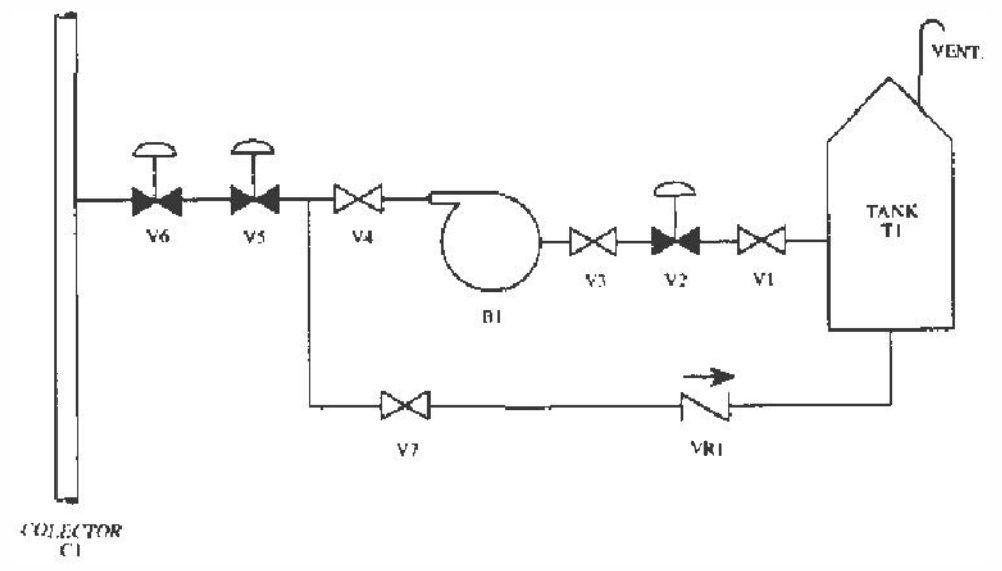

Figure 6: Simplified system diagram.

Table 1: Variables and associated physical meanings.

| Variable | Physical meaning |
|---|---|
| $ACA$ | Electric power failure |
| $B1A$ | Pump $B1$ fails to start |
| $B1B$ | Pump $B1$ fails after starting |
| $G1$ | Collector does not receive water flow |
| $G2$ | Valve $V6$ fails to open |
| $G3$ | Valve $V6$ does not receive water flow |
| $G4$ | Valve $V5$ fails to open |
| $G5$ | Valve $V5$ does not receive water flow |
| $G6$ | Valve $V4$ is closed |
| $G7$ | Valve $V4$ does not receive water flow |
| $G8$ | Pump $B1$ failure |
| $G9$ | Pump $B1$ does not receive water flow |
| $G10$ | Valve $V3$ is closed |
| $G11$ | Valve $V3$ does not receive water flow |
| $G12$ | Valve $V2$ fails to open |
| $G13$ | Valve $V2$ does not receive water flow |
| $G15$ | Valve $V1$ does not receive water flow |
| $M1$ | Element out of service (maintenance) |
| $O1$ | Valve $V4$ opened after maintenance |
| $O2$ | Valve $V3$ opened after maintenance |
| $SISA$ | Logic signal failure |
| $T1A$ | Tank failure |
| $T1B$ | Ventilation tank failure |
| $V1A$ | Valve $V1$ is blocked |
| $V2A$ | Valve $V2$ mechanical failure |
| $V2B$ | Valve $V2$ is blocked |
| $V3A$ | Valve $V3$ is blocked |
| $V4A$ | Valve $V4$ is blocked |
| $V5A$ | Valve $V5$ mechanical failure |
| $V5B$ | Valve $V5$ is blocked |
| $V6A$ | Valve $V6$ mechanical failure |
| $V6B$ | Valve $V6$ is blocked |

We are interested in the unavailability of the system. The operating policy is as follows:

1. The aim of the system is to supply water from tank $T1$ to collector $C1$.
2. The system must pump $B1$ and open the motorized valves $V2$, $V5$ and $V6$.
3. All valves are shown in Figure 6 in their normal positions (standby system).
4. Pump $B1$ is checked once a month; during the test, the pump is started and works for 10 minutes, making the water to flow through the manual valve $V7$ and the retention valve $VR1$, after opening the motorized valve $V2$ returning the water to tank $T1$.
5. The system works correctly even though valve $V7$ be open.
6. The maintenance to pump $B1$ is done once every five months. To this aim, valves $V3$ and $V4$ are closed and we assume that the sytem is unavailable if demand is required during the maintenance period, which takes 7 hours.
7. No other contributions to unavailability are considered.

8. In each recharge the availability of the system is checked.
9. Pump $B1$ must work for 24 hours in order to mitigate one accident.
10. The power supply of the pump and all motorized valves comes from train $A$, which has an estimated unavailability of $1.0 \times 10^{-3}$. Its failure probability during 24 hours is assumed neglegible.
11. Logical signals of the pump and motorized valves come from train $A$, which has an estimated unavailability of $1.0 \times 10^{-4}$.
12. Valves $V5$ and $V6$ are tested once a month.

In Figure 7 we show the graph associated with the Bayesian network, in which we have used a network avoiding replication of nodes, as it is usually done with fault tree diagrams, and showing the corresponding dependence structure.

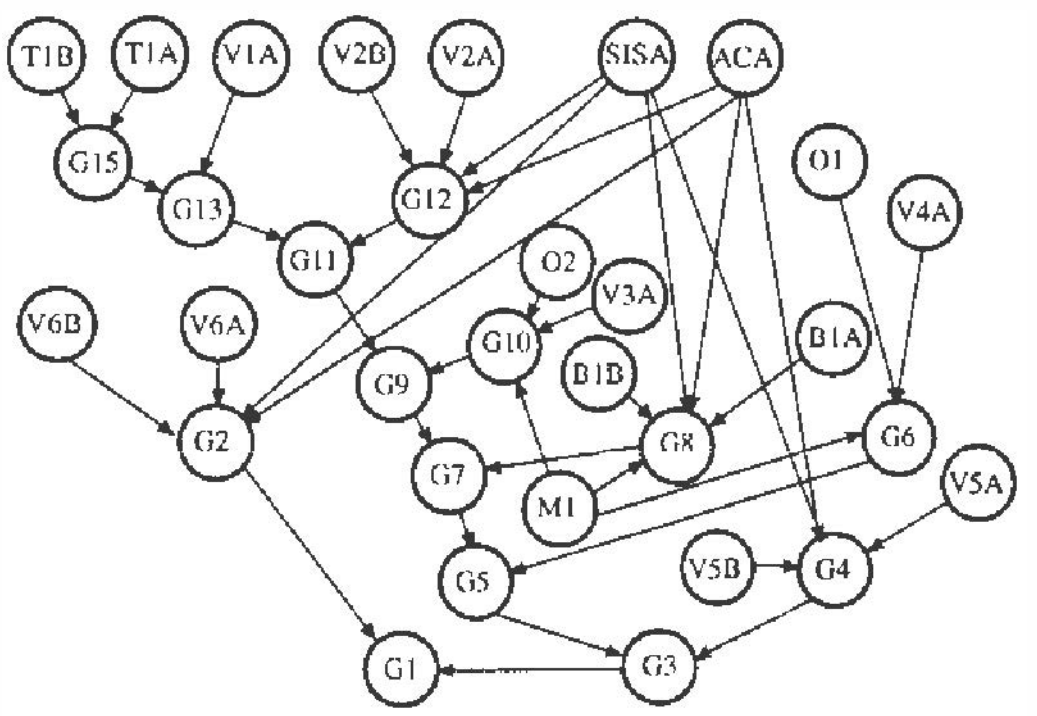

Figure 7: Directed acyclic graph.

We are interested in obtaining a confidence interval for the probability of failure of the system. In particular, we want to analyze the influence of the failure probabilities (unavailabilities) of the logic signal failure (SISA), the electric power system (ACA) and the maintenance policy (M1) on the probability of failure of the system. It can be shown that the probability of failure of the system can be written as

$$Z = P(G1 = 1) = h(x_1, x_2, x_3) = 1 - \alpha x_1 x_2 x_3, \quad (43)$$

where $x_1, x_2$ and $x_3$ are the probabilities of no failure associated with variables $SISA$, $ACA$ and $M1$, respectively, and $\alpha$ is a constant, which is close to 1 and depends on the failure probabilities of the remaining elements in the system.

Since Expression (43) involves the product $x_1 x_2 x_3$, we can use the results of the previous examples to this case, we can solve the sensitivity problem in two different ways:

**Exact Analysis** : Expression (43) shows that the cdf of $Z$ is

$$\begin{aligned} F_Z(z) &= Prob[Z \le z] = Prob[1 - \alpha x_1 x_2 x_3 \le z] \\ &= 1 - F_U[(1-z)/\alpha], \end{aligned} \quad (44)$$

where $F_U(u)$ is the cdf of $U = x_1 x_2 x_3$.

If we assume that $X_1, X_2$ and $X_3$ are iid uniform $U(\beta, 1)$ random variables, the cumulative distribution function of $U$ in the region $z < \beta^2$ is

$$\begin{aligned} F_U(u) = & \frac{-\beta^3 + u + u\log\beta + u(\log\beta)^2/2}{(1-\beta)^3} \\ & + \frac{u(\log u/\beta^2)^2/2 - (u + u\log\beta)\log(u/\beta^2)}{(1-\beta)^3} \end{aligned} \quad (45)$$

**Approximate Analysis** : It is clear that the minimum of $Z$ is $1 - \alpha$ and is attained at the point $(a_1, a_2, a_3) = (1, 1, 1)$. We can use:

**The No Rejection Method** : Using the techniques described in Section 2.1, if we simulate sequentially, $X_1, X_2$ and $X_3$, its simulation ranges become

$$L_1 = \beta \le X_1 \le \frac{\alpha\beta^3 + \epsilon}{\alpha\beta^2} = U_1 \quad (46)$$

$$L_2 = \beta \le X_2 \le \frac{\alpha\beta^3 + \epsilon}{\alpha x_1 \beta} = U_2 \quad (47)$$

$$L_3 = \beta \le X_3 \le \frac{\alpha\beta^3 + \epsilon}{\alpha x_1 x_2} = U_3. \quad (48)$$

Figure 8 shows the exact and the simulated tail for $\alpha = 0.999, \beta = 0.9999$ and $\epsilon = 0.00005$, corresponding to a sample size of $n = 1000$ and using the no rejection method.

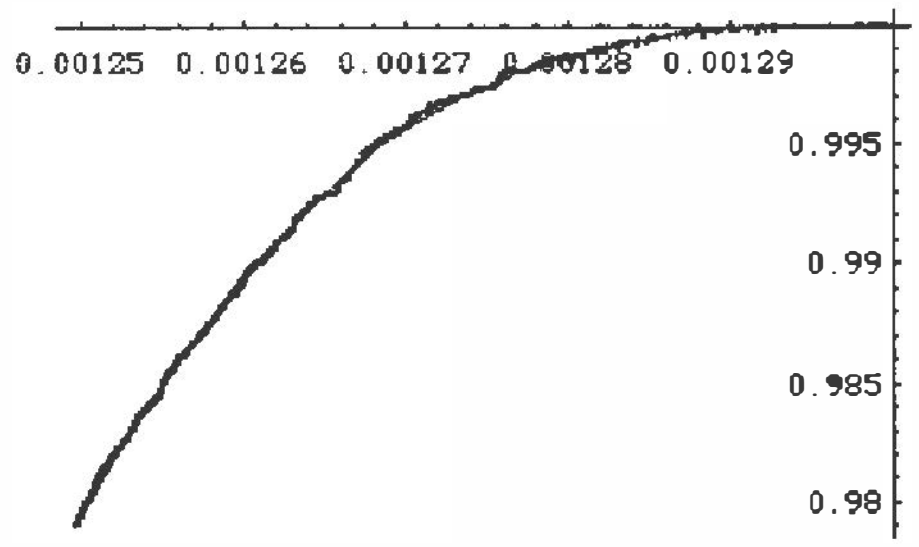

Figure 8: Exact and simulated tail of $h(x_1, x_2, x_3) = 1 - \alpha x_1 x_2 x_3$ for $\alpha = 0.999$ when $X_i, i = 1, 2, 3$ are uniforms $U(0.9999, 1)$ (No Rejection Method).

**The Reduced Rejection Method** : In this case, we cannot use the tangent hyperplane, since the simulation region does not lie on one side of it. Thus, we use the hyperplane passing through the points of intersection of the function $1 - \alpha x_1 x_2 x_3 = 1 - \alpha\beta^3 - \epsilon$ with the straight lines $\{x_1 = \beta, x_2 = \beta\}$, $\{x_1 = \beta, x_3 = \beta\}$ and $\{x_2 = \beta, x_3 = \beta\}$, which has the following equation

$$x_1 + x_2 + x_3 = 3\beta + \epsilon/(\alpha\beta^2). \quad (49)$$

and the corresponding score $w = \prod_{i=1}^{3} \frac{(U_i - L_i)}{(1-\beta)^3}$. From (49) we obtain

$$x_i = 3\beta + \epsilon/(\alpha\beta^2) - \sum_{j \ne i} x_j; i = 1, 2, 3 \quad (50)$$

Thus, if we simulate sequentially, $X_1, X_2$ and $X_3$, its simulation ranges become

$$L_1 = \beta \leq X_1 \leq \beta + \frac{\epsilon}{\alpha\beta^2} = U_1 \quad (51)$$

$$L_2 = \beta \leq X_2 \leq 2\beta - x_1 + \frac{\epsilon}{\alpha\beta^2} = U_2 \quad (52)$$

$$L_3 = \beta \leq X_3 \leq 3\beta - x_1 - x_2 + \frac{\epsilon}{\alpha\beta^2} = U_3. \quad (53)$$

and the corresponding score $w = \prod_{i=1}^{3} \frac{(U_i - L_i)}{(1-\beta)^3}$.

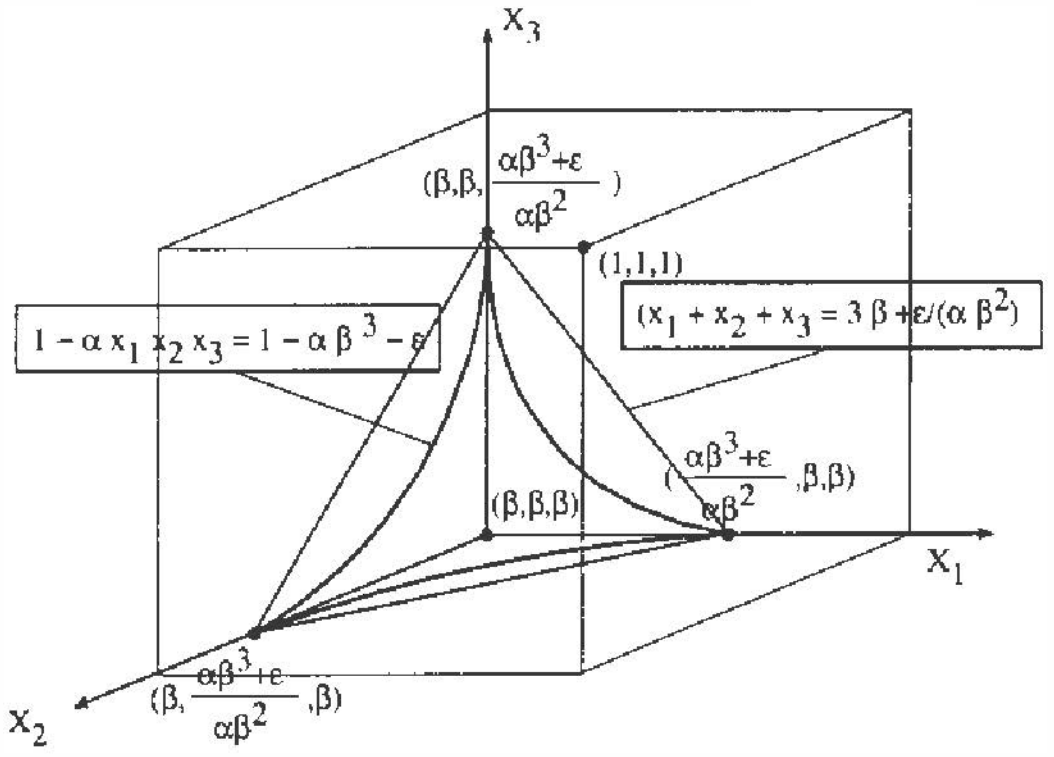

Figure 9: Illustration of the simulation region.

Figure 10 shows the exact and the simulated tail for $\alpha = 0.999, \beta = 0.9999$ and $\epsilon = 0.00005$, corresponding to a sample size of $n = 1000$ and a rejection proportion of $p = 0.001$. This allows us to obtain one-sided confidence intervals for the probability of failure of the system. For example, we can say that $(0, 0.00125)$ is a 0.98 confidence interval for the probability of failure of the system.

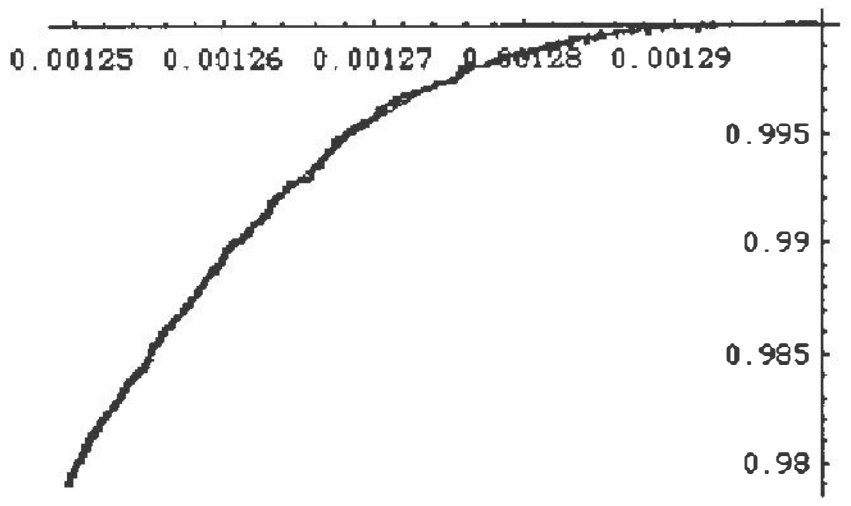

Figure 10: Exact and simulated tail of $h(x_1, x_2, x_3) = 1 - \alpha x_1 x_2 x_3$ for $\alpha = 0.999$ when $X_i, i = 1, 2, 3$ are uniforms $U(0.9999, 1)$ (Reduced Rejection Method).

## 4 CONCLUSIONS

Two efficient computational algorithms for simulating the left (right) tail of a random variable which is defined as an increasing (decreasing) invertible in each variable function of a set of basic variables has been given. One of the methods allows simulating the tail directly, i.e., all the simulated points are guaranteed to belong to the target tail; this leads to a good performance of the method. On the contrary, the proposed alternative method leads to a relatively low rejection proportion of simulated values when compared with the standard Monte Carlo method. Several theoretical examples and a real life example have been given to illustrate the method. Comparison of the real cdf in the tail with the simulated cdf shows that the method performs well especially if the simulation method is carefully selected for getting similar scores for all simulated instantiations. The method has immediate applications in many fields of reliability theory, as probability risk or security assessments of complex systems.

### Acknowledgements

We thank Iberdrola, the Leonardo Torres Quevedo Foundation of the University of Cantabria and Dirección General de Investigación Científica y Técnica (DGICYT) (project PB94-1056), for partial support of this work.